\documentclass{article}

\usepackage[dvipsnames]{xcolor}

\usepackage{natbib} %

\usepackage[utf8]{inputenc} %
\usepackage[T1]{fontenc}    %
\usepackage{hyperref}       %
\usepackage{booktabs}       %
\usepackage{graphicx}
\usepackage{duckuments}
\usepackage{placeins}
\usepackage{tcolorbox}

\usepackage{tikz}
\newcommand*\Circled[1]{\tikz[baseline=(char.base)]{
		\node[shape=circle,draw,inner sep=2pt] (char) {#1};}}
\newcommand*\RRectangle[1]{\tikz[baseline=(char.base)]{
		\node[shape=rectangle,rounded corners=2pt,draw,inner sep=2pt] (char) {#1};}}

\usepackage{natbib}
\usepackage{babel}

\usepackage{amsthm}
\usepackage{amssymb}
\usepackage{amsfonts} %
\usepackage{thmtools,thm-restate}
\usepackage{mathtools} %
\usepackage{mathcommand}
\usepackage{nicefrac}       %
\usepackage{bm}
\usepackage{etoolbox}
\usepackage{xspace}
\usepackage[colorinlistoftodos]{todonotes}

\usepackage{cleveref}

\usepackage{pifont}
\usepackage{graphicx}
\usepackage{booktabs} %
\usepackage{fancyhdr}
\usepackage{multirow}

\usepackage{url}            %
\usepackage{microtype}
\usepackage{subcaption} 

\usepackage{comment}

\excludecomment{draft}
\excludecomment{fullpaper}

\usepackage{enumitem}
\setitemize{noitemsep,topsep=0pt,parsep=0pt,partopsep=0pt}
\setenumerate{noitemsep,topsep=0pt,parsep=0pt,partopsep=0pt}

\graphicspath{{./figures/}}

\newtheorem{theorem}{Theorem}[section]

\newtheorem{definition}[theorem]{Definition}

\DeclareMathOperator{\opExpectation}{\mathbb{E}}
\newcommand{\E}[2]{\opExpectation_{#1} \left [ \ifblank{#2}{\:\cdot\:}{#2} \right ]}
\newcommand{\simpleE}[1]{\opExpectation_{#1}} %

\providecommand\given{\MidSymbol[\vert]}

\newcommand\MidSymbol[1][]{%
\nonscript\:#1
\allowbreak
\nonscript\:
\mathopen{}}

\DeclareMathOperator{\opInformationContent}{h}
\DeclarePairedDelimiterXPP{\ICof}[1]{\opInformationContent}{(}{)}{}{%
    \ifblank{#1}{\:\cdot\:}{#1}
}

\DeclareMathOperator{\opEntropy}{H}
\DeclarePairedDelimiterXPP{\Hof}[1]{\opEntropy}{[}{]}{}{%
    \renewcommand\given{\MidSymbol[\delimsize\vert]}
    \ifblank{#1}{\:\cdot\:}{#1}
}
\DeclarePairedDelimiterXPP{\xHof}[1]{\opEntropy}{(}{)}{}{%
    \ifblank{#1}{\:\cdot\:}{#1}
}

\DeclareMathOperator{\opMI}{I}
\DeclarePairedDelimiterXPP{\MIof}[1]{\opMI}{[}{]}{}{%
    \renewcommand\given{\MidSymbol[\delimsize\vert]}
    \ifblank{#1}{\:\cdot\:}{#1}
}

\DeclarePairedDelimiterXPP{\CrossEntropy}[2]{\opEntropy}{(}{)}{}{%
    \ifblank{#1#2}{\:\cdot\: \MidSymbol[\delimsize\Vert] \:\cdot\:}{#1 \MidSymbol[\delimsize\Vert] #2}
}

\DeclareMathOperator{\opKale}{D_\mathrm{KL}}
\DeclarePairedDelimiterXPP{\Kale}[2]{\opKale}{(}{)}{}{%
    \ifblank{#1#2}{\:\cdot\: \MidSymbol[\delimsize\Vert] \:\cdot\:}{#1 \MidSymbol[\delimsize\Vert] #2}
}

\DeclareMathOperator{\opp}{p}
\DeclarePairedDelimiterXPP{\pof}[1]{\opp}{(}{)}{}{%
    \renewcommand\given{\MidSymbol[\delimsize\vert]}
    \ifblank{#1}{\:\cdot\:}{#1}
}

\DeclarePairedDelimiterXPP{\pcof}[2]{\opp_{#1}}{(}{)}{}{%
    \renewcommand\given{\MidSymbol[\delimsize\vert]}
    \ifblank{#2}{\:\cdot\:}{#2}
}

\DeclarePairedDelimiterXPP{\hpcof}[2]{\hat{\opp}_{#1}}{(}{)}{}{%
    \renewcommand\given{\MidSymbol[\delimsize\vert]}
    \ifblank{#2}{\:\cdot\:}{#2}
}

\DeclareMathOperator{\opq}{q}
\DeclarePairedDelimiterXPP{\qof}[1]{\opq}{(}{)}{}{%
    \renewcommand\given{\MidSymbol[\delimsize\vert]}
    \ifblank{#1}{\:\cdot\:}{#1}
}

\DeclarePairedDelimiterXPP{\qcof}[2]{\opq_{#1}}{(}{)}{}{%
    \renewcommand\given{\MidSymbol[\delimsize\vert]}
    \ifblank{#2}{\:\cdot\:}{#2}
}

\DeclarePairedDelimiterXPP{\varHof}[2]{\opEntropy_{\ifblank{#1}{\:\cdot\:}{#1}}}{[}{]}{}{%
    \renewcommand\given{\MidSymbol[\delimsize\vert]}
    \ifblank{#2}{\:\cdot\:}{#2}
}

\DeclarePairedDelimiterXPP{\xvarHof}[2]{\opEntropy_{\ifblank{#1}{\:\cdot\:}{#1}}}{(}{)}{}{%
    \renewcommand\given{\MidSymbol[\delimsize\vert]}
    \ifblank{#2}{\:\cdot\:}{#2}
}

\newcommand{\Dtrain}{\boldsymbol{\mathcal{D}^\text{train}}}

\newcommand{\Dpool}{\boldsymbol{\mathcal{D}^\text{pool}}}
\newcommand{\Deval}{\boldsymbol{\mathcal{D}^\text{eval}}}

\newcommand{\ptest}[1]{\pcof{\text{test}}{#1}}

\newcommand{\ppool}[1]{\pcof{\text{pool}}{#1}}
\newcommand{\peval}[1]{\pcof{\text{eval}}{#1}}

\newcommand{\w}{\omega}
\newcommand{\W}{\Omega}

\newcommand{\xeval}{x^\text{eval}}
\newcommand{\xtest}{x^\text{test}}
\newcommand{\xtrain}{x^\text{train}}
\newcommand{\xacq}{x^\text{acq}}
\newcommand{\xpool}{x^\text{pool}}

\newcommand{\Xeval}{X^\text{eval}}
\newcommand{\Xtest}{X^\text{test}}
\newcommand{\Xtrain}{X^\text{train}}
\newcommand{\Xacq}{X^\text{acq}}
\newcommand{\Xpool}{X^\text{pool}}

\newcommand{\yeval}{y^\text{eval}}
\newcommand{\ytest}{y^\text{test}}
\newcommand{\ytrain}{y^\text{train}}
\newcommand{\yacq}{y^\text{acq}}
\newcommand{\ypool}{y^\text{pool}}

\newcommand{\Ytest}{Y^\text{test}}
\newcommand{\Ytrain}{Y^\text{train}}
\newcommand{\Yeval}{Y^\text{eval}}
\newcommand{\Yacq}{Y^\text{acq}}
\newcommand{\Ypool}{Y^\text{pool}}

\newcommand{\xevalset}{\{x^\text{eval}_i\}_i}
\newcommand{\xtestset}{\{x^\text{test}_i\}_i}
\newcommand{\xtrainset}{\{x^\text{train}_i\}_i}
\newcommand{\xacqset}{\{x^\text{acq}_i\}_i}
\newcommand{\xpoolset}{\{x^\text{pool}_i\}_i}

\newcommand{\Xevalset}{\{X^\text{eval}_i\}_i}
\newcommand{\Xtestset}{\{X^\text{test}_i\}_i}
\newcommand{\Xtrainset}{\{X^\text{train}_i\}_i}
\newcommand{\Xacqset}{\{X^\text{acq}_i\}_i}
\newcommand{\Xpoolset}{\{X^\text{pool}_i\}_i}

\newcommand{\yevalset}{\{y^\text{eval}_i\}_i}
\newcommand{\ytestset}{\{y^\text{test}_i\}_i}
\newcommand{\ytrainset}{\{y^\text{train}_i\}_i}
\newcommand{\yacqset}{\{y^\text{acq}_i\}_i}
\newcommand{\ypoolset}{\{y^\text{pool}_i\}_i}

\newcommand{\Ytestset}{\{Y^\text{test}_i\}_i}
\newcommand{\Ytrainset}{\{Y^\text{train}_i\}_i}
\newcommand{\Yevalset}{\{Y^\text{eval}_i\}_i}
\newcommand{\Yacqset}{\{Y^\text{acq}_i\}_i}
\newcommand{\Ypoolset}{\{Y^\text{pool}_i\}_i}

\newcommand{\xset}{\{x_i\}_i}

\newcommand{\yset}{\{y_i\}_i}

\newcommand{\xtrainsetfull}{\{x^\text{train}_i\}_{i \in \{1, \dots, |\Dtrain|\}}}
\newcommand{\xpoolsetfull}{\{x^\text{pool}_i\}_{i \in \{1, \dots, |\Dpool|\}}}

\newcommand{\ytrainsetfull}{\{y^\text{train}_i\}_{i \in \{1, \dots, |\Dtrain|\}}}

\newcommand{\xsetfull}{\{x_i\}_{i \in I}}

\newcommand{\ysetfull}{\{y_i\}_{i \in I}}

\newcommand{\EPIG}{EPIG\xspace}
\newcommand{\fullEPIG}{Expected Predictive Information Gain\xspace}
\newcommand{\EPIGBALD}{JEPIG\xspace}
\newcommand{\fullEPIGBALD}{Joint Expected Predictive Information Gain\xspace}

\newcommand\independent{\protect\mathpalette{\protect\independenT}{\perp}}
\def\independenT#1#2{\mathrel{\rlap{$#1#2$}\mkern2mu{#1#2}}}

\DeclareMathOperator*{\argmax}{arg\,max}

\newcommand{\tom}[1]{}
\newcommand{\andreas}[1]{}
\newcommand{\yarin}[1]{}

\begin{draft}
\newcommand{\btodo}[3]{\todo[author={#1}, color={#2},inline]{#3}}
\renewcommand{\andreas}[1]{\btodo{AK}{orange}{#1}}
\renewcommand{\tom}[1]{\btodo{TR}{SkyBlue}{#1}}
\renewcommand{\yarin}[1]{\btodo{YG}{Goldenrod}{#1}}
\end{draft}

\usepackage[accepted]{icml2021_arxiv}

\everypar{\looseness=-1}
\allowdisplaybreaks
\newcommand{\mytitle}{Test Distribution--Aware Active Learning: \\ A Principled Approach Against Distribution Shift and Outliers}
\icmltitlerunning{Test Distribution--Aware Active Learning}

\begin{document}

\twocolumn[
\icmltitle{\mytitle}

\icmlsetsymbol{equal}{*}

\begin{icmlauthorlist}
\icmlauthor{Andreas Kirsch}{oatml}
\icmlauthor{Tom Rainforth}{stats}
\icmlauthor{Yarin Gal}{oatml}
\end{icmlauthorlist}

\icmlaffiliation{oatml}{OATML, Department of Computer
Science,}
\icmlaffiliation{stats}{Department of Statistics, University of Oxford}

\icmlcorrespondingauthor{Andreas Kirsch}{andreas.kirsch@cs.ox.ac.uk}

\icmlkeywords{Machine Learning, ICML}

\vskip 0.3in
]

\printAffiliationsAndNotice{}  %

\begin{abstract}
    Expanding on \citet{mackay1992information}, we argue that conventional model-based methods for active learning---like BALD---have a fundamental shortfall: they fail to directly account for the test-time distribution of the input variables.
    This can lead to pathologies in the acquisition strategy, as what is maximally informative for model parameters may not be maximally informative for prediction:
    for example, when the data in the pool set is more disperse than that of the final prediction task, or when the distribution of pool and test samples differs.
    To correct this, we revisit an acquisition strategy that is based on maximizing the expected information gained about \emph{possible future predictions}, referring to this as the \fullEPIG (\EPIG).
    As \EPIG does not scale well for batch acquisition, we further examine an alternative strategy, a hybrid between BALD and \EPIG, which we call \fullEPIGBALD (\EPIGBALD).
    We consider using both for active learning with Bayesian neural networks on a variety of datasets, examining the behavior under distribution shift in the pool set.
\end{abstract}

\section{Introduction}
\label{sec:introduction}

Active learning provides a mechanism for effective training of machine learning models in settings where unlabelled data is plentiful but labelling is expensive~\citep{atlas1990training, settles2009active}.
It does so by carefully selecting which data points to acquire labels for, using information from previously acquired data to establish the points whose labels will be most informative for training.
This is most often done in a \emph{pool}-based setting, wherein we start with a large reservoir of unlabelled data points, known as the \emph{pool set} $\Dpool$, from which we sequentially choose points to label, after which they are removed from $\Dpool$ and added to the \emph{training dataset} $\Dtrain$ together with their acquired label.

The mechanism by which we choose points to label is known as an acquisition strategy and most commonly corresponds to choosing the data point which maximizes a prespecified \emph{acquisition function} that reflects the utility of acquiring a label for that point.
Strategies for constructing such an acquisition function can be based on principled information-theoretic considerations that allow formalizing the notion of the information that will be gained for labelling any given point.
These approaches usually require a probabilistic model $\pof{y \given x, \w}$ for label $y$ given input $x$, where $\w$ represents a realization of stochastic model parameters $\W$; a particularly common choice of model is a Bayesian neural network, wherein $\W$ represents the weights and biases.

Most notably, for $\W$ as a random variable with distribution given previous data $\pof{\w \given \Dtrain}$, we can define the \emph{expected information gain} (EIG) for $\W$ under $\pof{y \given x, \w}$~\citep{lindley1956measure,chaloner1995},
\begin{align}
\begin{split}
\text{EIG}(x) := \simpleE{\pof{y \given x,\Dtrain}}{} &\big[\xHof{\pof{\W \given \Dtrain}}
\\
&\phantom{\big[}-\xHof{\pof{\W \given y,x,\Dtrain}}\big],
    \end{split} \label{eq:original_EIG}
\end{align}
which is the expected reduction in entropy in $\W$ from observing the label of $x$, where
\mbox{$\pof{y \given x,\Dtrain} = \int \pof{y \given x,\w} \pof{\w \given \Dtrain} d\w$}
is the marginal predictive distribution of the model.
The EIG is also equal to the conditional mutual information between the parameters and the label, $\MIof{\W; Y \given x, \Dtrain}$, and is often referred to as \emph{BALD (Bayesian Active Learning by Disagreement)}~\citep{houlsby2011bayesian} when used with Bayesian neural networks \citep{neal1995bayesian} in the machine learning literature.

In this work, we highlight in \S\ref{sec:bald_shortfalls} that this framework can be sub-optimal when the goal is to learn a model with the \emph{best predictive performance} \citep{Roy2001TowardOA}.
Namely, in a machine learning context where we often care about predictions more than the parameter distribution,  $\W$ is not actually what we care about: it is merely a stepping stone to the predictions we will later make, with these (hypothetical) predictions being the ultimate quantity of interest.
This distinction can be surprisingly important: accurate prediction may not correspond to the most accurate knowledge of the model parameters themselves, particularly given the imperfections of commonly used models.
For example, in non-parametric settings it is possible to gain an infinite amount of information without learning anything about how to predict $Y \given x$ effectively in a region of interest if none of our training data are close by.
More generally, different information about the model parameters may not be equal in regards to how it enables effective prediction for a particular \emph{test-time distribution} over possible inputs $x$.

This is not a hypothetical setting. We may have data from different sources of varying fidelity and relevance to our problem. Indeed, if we have a large unlabelled dataset generated by scraping the internet, for example, we might have a significant degree of variation in how closely related individual data points are to the task we actually care about.
In an extreme case, the test-time distribution might consist of samples that cannot be directly labeled. This might apply in, for example, protein folding, where complex human proteins might make up the test-time distribution of interest, while the pool set contains simpler proteins which could be more easily crystallized and their  three-dimensional structure learnt.
    
\begin{figure}[t]
        \centering
        \includegraphics[width=\linewidth]{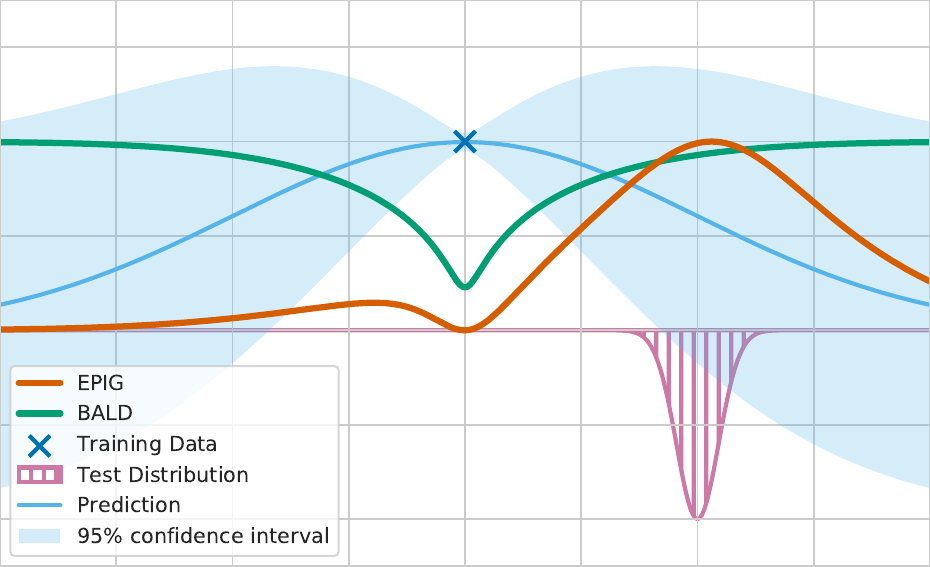}
        \vspace{-2em}
        \caption{High-level demonstration of the difference in the acquisition functions for Expected Information Gain (BALD) and \fullEPIG (EPIG).
        Here we consider a simple 1D regression task with a Gaussian process model, larger values of the acquisition function indicate it prefers acquiring at that point.
        We see that while BALD is predominantly concerned with labeling inputs far away from the data it has already seen,
        \EPIG also takes into account the need to sample in higher density regions of the test distribution.}
        \label{fig:toy_example}    
\end{figure}

To address this issue, we will strip Bayesian active learning back to its Bayesian experimental design foundations~\citep{lindley1956measure} and revisit in \S\ref{sec:epig} how we can reformulate acquisition functions in terms of the information gained about the potential predictions themselves \citep{mackay1992information}.
This is achieved by considering the prediction, $\Yeval$, at an arbitrary test input point $\Xeval$ as our quantity of interest, then marginalizing out over different possible test inputs $\Xeval$.
This leads to an acquisition function that we refer to as the \emph{\fullEPIG} (\EPIG) to stress that we care about the predictions unlike the regular EIG.
\EPIG can also be viewed as an EIG of its own, but where the quantity we wish to learn about is $\Yeval$ (instead of $\W$).
It was originally introduced as Mean Marginal Information Gain by \citet{mackay1992information}.
\begin{fullpaper}
We show how EPIG can be estimated effectively for a general class of problems and evaluate it empirically in setting of active learning for Bayesian neural networks on a variety of datasets.
\end{fullpaper}
A high-level demonstration of its behavior is given in~\Cref{fig:toy_example}.

Unfortunately, \EPIG can be quite expensive to estimate accurately, and thus optimize, in practice: while BALD can be estimated via a conventional Monte Carlo estimator for classification (but not regression) problems, \EPIG is always doubly intractable and therefore requires the use of expensive nested Monte Carlo estimators~\citep{rainforth2018nesting} or other strategies for nested estimation~\citep{foster2019variational}.
This can be particularly problematic in the batch acquisition setting, for which acquisition costs are already raised~\citep{kirsch2019batchbald}.
\andreas{however, we do not use a nested MC estimator for EPIG at the moment.}

To account for this, we also examine an alternative test distribution--aware acquisition function called the \emph{\fullEPIGBALD} (\EPIGBALD).
Like BALD, \EPIGBALD is still based around targeting information gain in the model parameters, but it discounts information that is not relevant to prediction on a given set of test samples.
Specifically, it removes from BALD any information that will remain unknown after we have acquired the labels for this given test set, on the basis that this information would not be helpful for prediction.
\citet{mackay1992information} referred to it as Joint Information Gain and prematurely dismissed it as it can be equivalent to BALD in certain cases that we will discuss in \S\ref{sec:epigbald}.
Its key distinction to \EPIG is that it allows us to introduce an approximation strategy that can reduce the cost of its optimization in the batch setting, increasing its practicality.

Finally, we discuss possible limitations of our approach in \S\ref{sec:limitations}, detail related work in \S\ref{sec:related work}, and present first results of an empirical evaluations using approxmiate BNNs in active learing in \S\ref{sec:experiments}.

To summarize, our key contributions are as follows:
\begin{itemize}
    \item we highlight the failure of conventional model--based active learning schemes to account for the test--time input distribution and subsequent issues this can cause;
    \item we examine \EPIG as an acquisition function that is explicitly geared to gaining information about test--time prediction;
    \item we examine \EPIGBALD as an alternative test-time input distribution--aware strategy that allows for cheaper acquisition of labels and thus practical application to batch settings.
\end{itemize}

\section{Background}
Throughout this paper, we follow the practical notation for information-theoretic quantities from \citet{kirsch2021practical} but also expand terms fully for clarity.
\textbf{Model Setup.} %
We treat the model parameters as a random variable $\W$ with prior
 distribution $\pof{\w}$. We denote the training set $\Dtrain=\{ ( \xtrain_i,\ytrain_i) \}_{1, \ldots, i \in |\Dtrain|}$, where $\xtrainsetfull$ are the input samples and $\ytrainsetfull$ the labels or targets.
The probabilistic model is 
\begin{math}
    \pof{y, \w | x} = \pof{y \given x, \w} \, \pof{\w},
\end{math}
where $y$ are outcomes of the random variable $Y$ denoting the label for input \(x\). 
Note here that the inputs are not treated as random variables in the model: we are focused on the typical active learning setting of purely discriminative, supervised, models.

To include multiple labels and inputs, we assume an exchangeable model and expand the model to joints of random variables $\ysetfull$ for given inputs $\xsetfull$ obtaining
\begin{align*}
    \pof{\yset, \w \given \xset}=\prod_{i \in I} \pof{y_i \given x_i, \w} \pof{\w}.
\end{align*}
The posterior parameter distribution $\pof{\w \given \Dtrain}$ is determined via Bayesian inference. We obtain $\pof{\w \given \Dtrain}$ using Bayes' theorem:
\begin{equation*}
    \pof{\w \given \Dtrain} \propto \pof{\ytrainset \given \xtrainset, \w} \pof{\w}.
\end{equation*}
which allows for predictions by marginalizing over $\W$:
\begin{equation*}
    \pof{y \given x, \Dtrain} = 
    \simpleE{\pof{\w \given \Dtrain}}{\pof{y \given x, \w}}.
\end{equation*}

\textbf{Bayesian Active Learning.} %
In pool-based active learning \citep{settles2009active}, we repeatedly select samples $\xacqset$ from an unlabelled pool set $\Dpool=\xpoolsetfull$ and ask an oracle for labels. We then add those newly labelled samples to the training set $\Dtrain$, retrain the model \mbox{$\pof{\w \given \Dtrain}$}. We repeat this process until the model satisfies our performance requirements.
Generally, samples are acquired in batches to avoid retraining the model all the time. However, we initially focus on individual acquisition using an acquisition function
\begin{math}
     a(x, \pof{\w \given \Dtrain})
\end{math}:
\begin{align*}
    \xacq = \argmax_x a(x, \pof{\w \given \Dtrain}).
\end{align*}
As explained in the introduction BALD was introduced as a one-sample acquisition function of the expected information gain between the prediction $\Yacq$ for an input $\xacq$ and the stochastic model parameters $\W$.
Rewriting~\cref{eq:original_EIG} in our notation for Bayesian active learning, we get
\begin{align}
    a_\text{BALD}(\xacq, \pof{\w \given \Dtrain}) \coloneqq \MIof{\Yacq; \W \given \xacq, \Dtrain},
\end{align}
where we implicitly marginalize over the parameters \(\W\).

\section{Shortfalls with BALD: The Need to Make Acquisition Test-Time Distribution--Aware}
\label{sec:bald_shortfalls}

In this work we highlight, and look to address, a shortfall of BALD in this supervised discriminative learning context: it fails to account for the test--time distribution of the inputs $x$.
In fact, BALD does not consider the inputs as a random variable at all: $a_\text{BALD}$ is a function of the input and the pool set represents only a finite set of points we wish to optimize this function over.

In some circumstances, this can be advantageous.
For example,~\citet{farquhar_statistical_2020} showed that much of the practical success of active learning stems not from improving training per se, but from implicitly biasing towards models which tend to have better generalization properties than those trained by empirical risk minimization.

However, this will not always be the case: there is always, at least implicitly, some \emph{test-time} distribution over the inputs that we care about making predictions for.
If we fail to take this into account, there is no guarantee that the learned model will be well-suited to the prediction task we actually care about: the entropy of the posterior over our model parameters need not be representative of our predictive uncertainty for a particular test-time input distribution.

When we have data from different sources of varying fidelity and relevance to our problem, or the test-time distribution cannot be labelled at all---as described in \S\ref{sec:introduction}---%
we may be unsure upfront which data might be most useful so want to use a generic model and allow access to the whole pool set, but here BALD is likely to choose samples that are not sufficiently focused on the task at hand.
Since it does not take into account any notion of a test-time distribution over $x$, it is unlikely to focus on the region of interest in the manner we desire.
In fact, it can have a tendency to pick out `unusual' samples, which is likely to be counterproductive in this context \citep{karamcheti2021mind}.

This particular scenario is analogous to a train-test distribution shift in conventional machine learning settings.
However, the problems with BALD are more deep rooted than this as it does not even have a notion of a pool set distribution in the first place. 
Thus we need not have a `distribution shift,' just a test-time input distribution that is important to take account of.

\andreas{NOTE: Increasing the size of the pool set is not enough: we need to increase the diversity of the pool set. 
The failure case we describe here only becomes an issue for batch acquisition. Individual acquisition will sample one "duplicated" outlier from the pool set and then go on looking for the next, so we need to ensure that there are lots of different outliers. Ie stupid case with a non-parametric model: our test distribution is drawn from $[0,1]$ but there is a repeated outlier sample at $-10$ in the pool set: BALD will only pick the outlier once or twice and that's it (until it has acquired samples from all other regions at least once).
we might want to look at the "dimensionality" of the pool distribution in comparison to the test distribution (taking into account the "length scale" of our model) to construct cases where we can easily have many more different outliers than interesting samples.

\textbf{The curse of dimensionality might be important to mention here} because usually in DNNs are inputs are bounded, and hence the argument of going towards +-infinity is not as strong?
}
This is perhaps most easily seen by considering what happens when we increase the size of the pool set.
Adding samples to the pool will typically increase, and cannot decrease, its diversity and, in particular, the number of unrepresentative and extreme samples it contains.
BALD has a tendency to actively pick these samples---as they often have high uncertainty due to being far from other points in the data---and is therefore liable to suffer from diminishing performance as the representative samples required for prediction become rarer in the acquired data.
This problem can become even more pronounced if the pool set contains outliers, with the total number of outliers naturally increasing as the pool set is enlarged.

To give a more concrete demonstration for this surprising behavior that larger pool sets can actually harm the performance of BALD, consider the scenario shown in \Cref{fig:toy_example}.
Here we see that the BALD scores increase away from the already selected training data and ignore the test-time distribution; the true optima of the BALD objective for $x\in\mathbb{R}$ actually occurs at $x=\pm \infty$.
If we now assume that samples from the pool are drawn independently from some unbounded underlying distribution $p_{\text{pool}}(x)$, then increasing the size of the pool set will lead to BALD choosing points further and further from our actual region of interest, potentially significantly degrading performance.

Another important failure mode of BALD can occur when our model is significantly overparameterized.
For example, we might have a non-parametric model where the points acquired will be used directly in the predictions and so it is essential to enforce a congruence between the two.
We might also have a parametric model where different parameters have different degrees of influence on predictive capabilities.
For example, we may have a hierarchical model where the hyperparameters are of particular importance, or we might have nuisance parameters that will be of limited help for prediction.
In summary, BALD can lead to undesirable behavior whenever the joint entropy of the parameters of the model is not a sufficiently accurate measure of predictive capability.

\begin{figure}[t]
    \begin{tcolorbox}[colback=white,colbacktitle=white,coltitle=black,fonttitle=\bfseries,title=Expected Reduction in Generalization Loss, parbox, left=2mm, right=2mm]
    {
        \renewcommand{\xacqset}{\xacq}
        \renewcommand{\Yacqset}{\Yacq}
        \renewcommand{\yacqset}{\yacq}
        \EPIG measures the expected reduction in the uncertainty in a model's predictions on test points:
        \begin{align*}
            &\MIof{\Yeval; \Yacqset \given \Xeval, \xacqset, \Dtrain} =\\
            &\quad \Hof{\Yeval \given \Xeval, \Dtrain} \\
            &\quad \quad - \Hof{\Yeval \given \Xeval, \Yacqset, \xacqset, \Dtrain}. 
        \end{align*}
        Importantly, this objective is equivalent to minimizing the \emph{expected generalization loss} under the model's predictions:
        \begin{align*}
            &\Hof{\Yeval \given \Xeval, \Yacqset, \xacqset, \Dtrain} = \\
            &\quad = \simpleE{\peval{\xeval}} \\
            &\quad \quad \simpleE{\pof{\yacqset \given \xacqset, \Dtrain}} \\
            &\quad \quad \simpleE{\pof{\yeval \given \xeval, \yacqset, \xacqset, \Dtrain}}\mathcal{L}(\xeval, \yeval),
        \end{align*}
        with 
        \begin{align*}
            &\mathcal{L}(\xeval, \yeval) \coloneqq -\log \pof{\yeval \given \xeval, \yacqset, \xacqset, \Dtrain},
        \end{align*}
        where we have marginalized over the model parameters
        \begin{align*}
            &\w \sim \pof{\w \given \yacqset, \xacqset, \Dtrain}.   
        \end{align*}
        That is, the same samples maximize EPIG and minimize the expected generalization loss \citep{Roy2001TowardOA}.
    }
    \end{tcolorbox}
\end{figure}

\section{Method}
Here, we revisit the \emph{\fullEPIG (\EPIG)} acquisition function for individual acquisition, provide intuitions, and show how it can be computed. \EPIG was originally introduced as the Mean Marginal Information Gain in \citet{mackay1992information}, and we also examine its connection to the expected error reduction acquisition function in non-Bayesian active learning introduced in \citet{Roy2001TowardOA}. We consider the batch acquisition setting in the next section, in which we introduce \EPIGBALD as an alternative to \EPIG.

\subsection{Expected Predicted Information Gain}
\label{sec:epig}
\textbf{Evaluation Samples.} To be able to gain knowledge about the test distribution, we assume that we have access to the unlabelled test distribution, that is we can draw unlabelled \emph{evaluation samples} \(\xeval \sim \ptest{\xeval}\).
Similar to other unlabeled samples in active learning, these samples could be either provided in a stream-based setting where we can repeatedly draw new unlabelled samples i.i.d or in a pool-based setting where we only have a fixed reservoir of evaluation samples, which we call the \emph{evaluation set} \(\Deval\).
In the pool-based setting, we denote the \emph{empirical} distribution of the evaluation set by \(\peval{\xeval}\), whereas in the stream-based setting, we have \(\peval{\xeval} \coloneqq \ptest{\xeval}\). this allows us to abstract away the exact setting in the following exposition.

\textbf{Objective.} The overall objective is to minimize the population risk, also sometimes referred to as the generalization loss, which we assume here to use the common cross-entropy loss between the test distribution and a model's predictive distribution:
\begin{align}
&\CrossEntropy{\ptest{\Ytest, \Xtest}}{\pof{\Ytest \given \Xtest}} \coloneqq \notag \\
&\quad =\E{\ptest{\ytest, \xtest}} {-\log \pof{\ytest \given \xtest}}.
\end{align}
We want to construct an acquisition function which is conducive towards this goal. We will show that \EPIG does just that.

\begin{definition}
    The \emph{\fullEPIG (\EPIG)} is the expectation of the expected information gain for the prediction $\Yeval$ over samples $\xeval \sim \peval{\xeval}$ given the predictions $\Yacq$ for the candidate sample $\xacq$:
    \begin{align}
        & \MIof{\Yeval ; \Yacq \given \Xeval, \xacq, \Dtrain} \\
        &= \simpleE{\peval{\xeval}} \MIof{\Yeval ; \Yacq \given \xeval, \xacq, \Dtrain} \\
        &= \simpleE{\peval{\xeval}} \simpleE{\pof{\yeval, \yacq \given \xeval, \xacq, \Dtrain}} \label{eq:compute_EPIG}\\
        & \quad \left [
        -\log \frac{\pof{\yeval \given \xeval, \Dtrain}\pof{\yacq \given \xacq, \Dtrain}}{\pof{\yeval, \yacq \given \xeval, \xacq, \Dtrain}} \right ]. \notag
    \end{align}
\end{definition}
We can directly relate this to the objective we have specified above as detailed in \RRectangle{\textbf{Expected Reduction in Generalization Loss}}: \EPIG aims to reduce the expected generalization loss given the model's own predictions.

\textbf{Evaluation of \EPIG.} %
We can evaluate \EPIG using the posterior distribution $\pof{\w \given \Dtrain}$. 
That is, for specific \(\xeval, \xacq\), we use the joint density
\(\pof{\yeval, \yacq \given \xeval, \xacq, \Dtrain} = \simpleE{\pof{\w \given \Dtrain}}{\pof{\yeval \given \xeval, \w} \, \pof{\yacq \given \xacq, \w} d\w}\) that we obtain by marginalizing over the model parameters \(\pof{\w \given \Dtrain}\) to obtain \(\pof{\yeval \given \xeval \Dtrain}\) and \(\pof{\yacq \given \xacq, \Dtrain}\) by marginalizing over \(\yeval\) and \(\yacq\), respectively. These marginalizations are exact because we can enumerate all \(y\) in a classification task. 
By sampling $M$ model parameter samples \(\w_i \sim \pof{\w \given \Dtrain}\), we obtain a standard Monte-Carlo estimator using the deterministic transformations yielded by these marginalizations and the function \(-p \log p\) using the Monte-Carlo estimate \(\frac{1}{M} \sum_{i=1}^M \pof{\yeval \given \xeval, \w_i} \, \pof{\yacq \given \xacq, \w_i}\) for the joint density. Finally, we naively sample evaluation samples \(\xeval\) independently and obtain an estimate for \EPIG (or in the case of a finite evaluation set we can iterate over all samples).

\subsection{\fullEPIGBALD and Batch Acquisition}
\label{sec:epigbald}
\textbf{Batch Acquisition.} %
In practice, samples are acquired in batches to avoid retraining the model for each acquired sample. We score possible candidate batches $\xacq$ of \emph{acquisition batch size} $B$ using an acquisition function $a(\xacqset, \pof{\w \given \Dtrain})$ and pick the highest scoring batch:
\begin{equation*}
    \underset{\{\xacq_i\}_{i \in\{1,\ldots,B\}} \subseteq \Dpool}{\argmax} a(\xacqset, \pof{\w \given \Dtrain}).
\end{equation*}
BALD was introduced as a single-sample acquisition function of the expected information gain between the prediction $\Yacq$ for an input $\xacq$ and the stochastic parameters $\W$:
\begin{math}
    \MIof{\Yacq; \W \given \xacq, \Dtrain}.
\end{math}
This was trivially extended to batch acquisition by selecting the top-k highest scorers as a batch \citep{gal2017deep}. 
In \citet{kirsch2019batchbald}, this approach was shown to lead to the selection of redundant samples, and instead the one-sample case was canonically extended to the batch acquisition case using the expected information gain between the \emph{joint} of the predictions $\Yacqset$ for the batch candidates $\xacqset$ and the model parameters $\W$ (\emph{BatchBALD}):
\begin{multline*}
    a_\text{BatchBALD}(\xacqset, \pof{\W \given \Dtrain}) \coloneqq \\
    = \MIof{\Yacqset; \W \given \xacqset, \Dtrain} \\
    = \MIof{ \Yacq_1, \ldots, \Yacq_{b}; \W \given \xacq_1, \ldots, \xacq_{b}, \Dtrain}.
\end{multline*}
In practice, samples in a batch are selected greedily in \citet{kirsch2019batchbald} because this expected information gain is submodular, and thus greedy selection is $1-\tfrac{1}{e}$ optimal \citep{krause2014submodular}.

\textbf{Notation.}
Instead of $\Yevalset$, $\xevalset$, we will write $\boldsymbol{\Yeval}$, $\boldsymbol{\xeval}$ and so on to to save space as, canonically, all terms can be extended to sets by substituting the random variables for the product spaces.
Also note that lower-case variables like $\yeval$ are outcomes and upper-case variables like $\Yeval$ are random variables, with the exception of the datasets $\Dpool, \Dtrain,$ etc, which are sets of outcomes.

{
\renewcommand{\xevalset}{\boldsymbol{\xeval}}
\renewcommand{\xtestset}{\boldsymbol{\xtest}}
\renewcommand{\xtrainset}{\boldsymbol{\xtrain}}
\renewcommand{\xacqset}{\boldsymbol{\xacq}}
\renewcommand{\xpoolset}{\boldsymbol{\xpool}}

\renewcommand{\Xevalset}{\boldsymbol{\Xeval}}
\renewcommand{\Xtestset}{\boldsymbol{\Xtest}}
\renewcommand{\Xtrainset}{\boldsymbol{\Xtrain}}
\renewcommand{\Xacqset}{\boldsymbol{\Xacq}}
\renewcommand{\Xpoolset}{\boldsymbol{\Xpool}}

\renewcommandPIE{\yevalset}{\boldsymbol{y}^{\text{eval}\GetExponent{#3}}}
\renewcommand{\ytestset}{\boldsymbol{\ytest}}
\renewcommand{\ytrainset}{\boldsymbol{\ytrain}}
\renewcommand{\yacqset}{\boldsymbol{\yacq}}
\renewcommand{\ypoolset}{\boldsymbol{\ypool}}

\renewcommand{\Ytestset}{\boldsymbol{\Ytest}}
\renewcommand{\Ytrainset}{\boldsymbol{\Ytrain}}
\renewcommand{\Yevalset}{\boldsymbol{\Yeval}}
\renewcommand{\Yacqset}{\boldsymbol{\Yacq}}
\renewcommand{\Ypoolset}{\boldsymbol{\Ypool}}

\begin{figure}[t!]
    \begin{tcolorbox}[colback=white,colbacktitle=white,coltitle=black,fonttitle=\bfseries,title=Expected Reduction of Epistemic Uncertainty, parbox, left=2mm, right=2mm]
        Since BALD is known to measure epistemic uncertainty \citep{smith2018understanding}, we can take another look at the two mutual information terms:
        Combining above perspectives, we can examine \EPIGBALD once more:
        \begin{align}
            & \MIof{\Yevalset; \Yacqset ; \W \given \xacqset, \xevalset, \Dtrain} = \notag \\
            & \quad  = \MIof{\Yacqset ; \W \given \xacqset, \Dtrain} \quad \Circled{5}
            \label{eq:EREU} \\
            & \quad \quad - \MIof{\Yacqset ; \W \given \xacqset, \Yevalset,\xevalset, \Dtrain} \quad \Circled{6}. \notag
        \end{align}    
        Intuitively, the first term \Circled{5} is large when the model has high epistemic uncertainty about its prediction at $\xacqset$, and learning the true label would thus be informative for the model parameters,
        while the second term \Circled{6} captures the epistemic uncertainty about the model's prediction at $\xacqset$ assuming we had obtained labels for evaluation samples.
        This second term is small when $x$ is similar to drawn evaluation samples and the model can explain it well given the pseudo-labels. 
        Thus, the \EPIGBALD score is large when the first term is large and the second term is small, so learning $\xacqset$ is informative for the model and $\xacqset$ is similar to evaluation samples. 
        In reverse, the \EPIGBALD score will be small, when knowing about evaluation samples makes no difference for the epistemic uncertainty of the acquisition candidates, which means that they are unrelated.
    \end{tcolorbox}
\end{figure}

\textbf{Batch Acquisition using \EPIG.}
While the expectation in \EPIG can evaluated for individual acquisition, the batch setting is computationally more complex. Following  \citet{kirsch2019batchbald}, for batch acquisition, we need to maximize
\begin{math}
    \MIof{\Yeval ; \Yacqset \given \Xeval, \xacqset, \Dtrain}.
\end{math}
Unlike the expected information gain, this term is not submodular. However, as the global subset problem is not feasible, we will examine the case of using greedy selection nonetheless.
Computing \(\MIof{\Yeval ; \Yacqset \given \Xeval, \xacqset, \Dtrain}\) requires estimating a joint density in \(i+1\) many variables for each $\xeval$ sample for the \(i\)-th element in the acquisition batch:
\begin{math}
    \pof{\yeval, \yacqset \given \xeval, \xacqset}.
\end{math}
Overall, for an acquisition batch of size \(B\), this requires $O(\lvert \Deval \rvert \, \lvert \Dpool \rvert^B)$ many joint densities with 2 to $B+1$ many variables compared to $O(\lvert \Dpool \rvert^B)$ for BatchBALD with 1 to $B$ many variables.
Unfortunately, BatchBALD has already been found to be computationally intractable for larger acquisition batch sizes in practice \citep{kirsch2019batchbald}, and here we consider an additional variable by default, while we also have to evaluate this term for many different $\xeval$.

\textbf{Joint \fullEPIG.} %
Instead, we can examine the related objective
\begin{math}
\MIof{\Yevalset; \Yacqset \given \xacqset, \xevalset, \Dtrain},
\end{math}
which conditions on multiple evaluation set samples instead of taking an expectation over these samples, where \(\xevalset \sim \peval{\xevalset}\).

Furthermore, by extending the mutual information to a triple mutual information which takes into account the model parameters we can rewrite this expression as a difference between two (Batch-)BALD terms, which we denote \fullEPIGBALD (\EPIGBALD):
\begin{align*}
    & \MIof{\Yevalset; \Yacqset \given \xacqset, \xevalset, \Dtrain} = \\
    & \quad = \MIof{\Yevalset; \Yacqset ; \W \given \xacqset, \xevalset, \Dtrain} \\
    & \quad  = \MIof{\Yacqset ; \W \given \xacqset, \Dtrain} \\
    & \quad \quad - \MIof{\Yacqset ; \W \given \xacqset, \Yevalset,\xevalset, \Dtrain}.
\end{align*}
This allows for more efficient evaluation using approximate Bayesian neural networks as we detail in \S\ref{sec:practical_approximation} in the appendix. As explained in \RRectangle{\textbf{Expected Reduction of Epistemic Uncertainty}}, \EPIGBALD also has an intuitive explanation as expected reduction in epistemic uncertainty when taking into account evaluation samples.

\textbf{Relation of \EPIGBALD to BALD.} 
From \cref{eq:EREU} from the intution box for \EPIGBALD, we see that we have
\begin{align}
& \MIof{\Yevalset; \Yacqset ; \W \given \xacqset, \xevalset, \Dtrain} \notag \\
& \quad \le \MIof{\Yacqset ; \W \given \xacqset, \Dtrain},
\end{align}
that is, \EPIGBALD is upper-bounded by BALD---a two-term mutual information is always non-negative---and \EPIGBALD is equivalent to BALD exactly when \(\MIof{\Yacqset ; \W \given \xacqset, \Yevalset,\xevalset, \Dtrain}\) is zero. 
This is the case when the distribution of \(\Yacqset \given \xacqset\) is fully predicted by any \(\yevalset \given \xevalset \sim \pof{\yevalset \given \xevalset, \Dtrain}\), independently of the posterior \(\pof{\w \given \yevalset, \xevalset, \Dtrain}\).
Qualitatively, BALD and \EPIGBALD will trivially be equal when $\xacqset \subseteq \xevalset$, while in the case of non-parametric models, this will trivially not be the case when $\xacqset$ is not ``near'' the evaluation set. Intuitively, \EPIGBALD is different from BALD exactly when BALD fails: for outlier pool samples which are not similar to the test--time distribution, in which case \EPIGBALD will tend towards 0 as the two terms cancel out.

Note that \citet{mackay1992information} analyzed a similar acquisition function, which he called Joint Information Gain, yet dismissed it because in the specific context of that work, using Bayesian linear regression with simple homoscedastic noise, it is equivalent to BALD.

\textbf{Relation of \EPIGBALD to \EPIG.} %
In the following, we will assume that we have a finite amount of evaluation samples in an evaluation set for simplicity.
\begin{restatable}{proposition}{relationjpigepig}
    Maximizing \(\MIof{\Yeval; \Yacqset \given \Xeval, \xacqset, \Dtrain}\) is equivalent to minimizing \(\Hof{\Yeval \given \Xeval, \Yacqset, \xacqset, \Dtrain}\) which upper-bounds \(\frac{1}{\lvert \Deval \rvert} \Hof{\Yevalset \given \xevalset, \Yacqset, \xacqset, \Dtrain}\) whose minimization is equivalent to maximizing \(\frac{1}{\lvert \Deval \rvert} \MIof{\Yevalset; \Yacqset \given \xevalset, \xacqset, \Dtrain}\).
\end{restatable}
Hence, maximizing \EPIG also maximizes \EPIGBALD, but not the other way around, which is exactly the case when \EPIGBALD performs like BALD.
\andreas{this is a bit of a difficult argument but the intuition is that \EPIGBALD can minimize the joint predictive uncertainty by simply compressing the model better (and thus creating larger correlations), which might help generalization, but does not effect the actual predictive uncertainty in itself...}

\textbf{Evaluation of \EPIGBALD.}
We could maximize $\MIof{\Yevalset; \Yacq \given \xevalset, \xacq, \Dtrain}$ via
\begin{align}
    &\MIof{\Yevalset; \Yacq \given \xevalset, \xacq, \Dtrain} = \notag \\
    & \quad =
    \Hof{\Yacq \given \xacq, \Dtrain} \notag \\
    & \quad \quad - \Hof{\Yacq \given \xacq, \Yevalset, \xevalset, \Dtrain}.
\end{align}
Note that the second term is expensive to evaluate: 
\begin{align*}
    & \Hof{\Yacqset \given \xacqset, \Yevalset, \xevalset, \Dtrain} = \notag \\
    & \quad = \simpleE{\pof{\yevalset \given \xevalset, \Dtrain}} \Hof{\Yacqset \given \xacqset, \yevalset, \xevalset, \Dtrain}. \label{eq:pig_pseudolabel_conditional}
\end{align*}
That is, we need to train a model on $\Dtrain$ then sample joint label predictions $\yevalset$ for the evaluation set $\xevalset$ using the model, and then, for each such sampled joint prediction, we evaluate the conditional entropy $\Hof{\Yacqset \given \xacqset, \W'}$ using new $\w' \sim \pof{\w \given \yevalset, \xevalset, \Dtrain}$, which requires additional training using the labeled training set augmented with the sampled labels \(\Dtrain \cup \{\yeval_i, \xeval_i\}_i\).

In appendix \S\ref{sec:practical_approximation}, we present a computationally tractable approximation for this term using `\emph{self-distillation}`, where we train a new model on predictions for evaluation samples from the current model trained on $\Dtrain$. On a high-level, this yields a model that fulfills the intuitions presented in \RRectangle{\textbf{Expected Reduction of Epistemic Uncertainty}}. We refer to the appendix for more details.

\section{Limitations}
\label{sec:limitations}
While \EPIG can only be evaluated efficiently when using few evaluation samples, \EPIGBALD with self-distillation is an approximation that is also well motivated and potentially faster to evaluate in the batch setting.
However, it requires training an additional model. 
This could be sped up by using warm-starting \citep{ash2019warmstarting}.
Furthermore, when the labels are used, that is, we have a validation set instead of an evaluation set, the model can be trained only once assuming the validation set is sufficiently large. We explore this further in \citet{mindermann2021prioritized}.
At the same time, two (Batch)BALD terms need to be computed for \EPIGBALD which can be slow and does not scale well beyond small acquisition batch sizes.
Thus, \EPIG and \EPIGBALD represent a trade-off between the time it takes to compute the expectation over evaluation samples and training another model. In the case of individual acquisition, \EPIG is strongly favored. In the case of batch acquisition, \EPIGBALD might be faster. 
\citet{powerbald} suggest a simple method to avoid computing BatchBALD terms, trading off precision for performance. We do not examine this approach in detail but only present a first result in the experiment section.

Unlike BALD, \EPIG is not submodular, and thus greedy acquisition is not guaranteed to obtain $1-\tfrac{1}{e}$ optimality for batch acquisition, even though we have not experienced any degredation in comparison with BALD/BatchBALD empirically. Indeed, \EPIGBALD performs on par or better than BALD in the regular active learning case without distribution shift, too. 
At the same time, neither BALD nor \EPIG are adaptive submodular, and statements about global optimality cannot be made \citep{golovin2017adaptive, fosterThesis} which has not been an issue in practice.

Moreoever, \citet{ash2021gone} recently introduced a new forward-backward strategy that improved performance for a non-submodular objective. We hypothesize it might also further improve performance of \EPIG and \EPIGBALD, but we leave this to future work.

Finally, we present two scenarios where \EPIG might not perform better than BALD:
Firstly, if a task's performance is dominated by a more general sub-task, there might be many samples which are not part of the test--time distribution yet still highly informative for the more general sub-task. For image data, this can be the case when feature learning is of particular importance, and convolution kernels can be learnt from image data, no matter the label or overall task. In this regime, both BALD and \EPIG will perform similarly, yet will likely outperform random acquisition.
Secondly, if the model's architecture and its inductive biases have specifically evolved for a task, there will be a high overlap between the model parameters and the task's test--time input distribution as model architectures which converge faster are preferred as prior research artifacts. BALD and \EPIG might perform similarly in this case, too.
\andreas{the insights from this are: maybe most computer vision tasks are too `easy` because many samples will be informative for learning good feature kernels at first, in particular samples that do not correspond to a specific class for example. The confounder for EPIG vs BALD is that we need to find tasks where the dataset will contain lots of truly useless samples for the task. Ie if we had lots of game screenshots in RL and we only want to learn how to play a single game instead of something else. and we do not learn from pixels but from states directly, bc initially mostly any screenshot will be helpful for the model so will not help EPIG and BALD separate.
similarly, we need a model with enough capacity to learn a lot of different stuff and/or be a mixture such the model parameters and the test distribution won't be too correlated. maybe we should not train on CIFAR-10 but something larger like ImageNet or so and the task ought to be figuring out different subclasses of dogs or so and we do not do OOD in any way but strong class imbalance and start with a pretrained model (except for the last layer?)}

\section{Related Work}
\label{sec:related work}
The Expected Information Gain was introduced in Bayesian optimal experiment design by \citet{lindley1956measure}. 
While Expected Information Gain focuses on the formulation of the mutual information term as a reduction in model posterior uncertainty given a potential sample, in active learning, BALD was introduced as an equivalent term which focuses on the predictive disagreement \citep{houlsby2011bayesian}.
For high-dimensional data, BALD has been applied to Bayesian deep learning models using Monte-Carlo dropout \citep{gal2017deep}. BALD was further extended to BatchBALD to correctly capture redundancies in the batch acquisition setting \citep{kirsch2019batchbald}.

\citet{mackay1992information} introduces the Total Information Gain and Mean Marginal Information Gain, which correspond to the Expected Information Gain and \fullEPIG, respectively, and examines them using Bayesian linear regression \emph{specifically}.
In this work, MacKay also examines the Joint Information Gain, which corresponds to the two term version of \EPIGBALD, and shows that it is equivalent to BALD when assuming constant aleatoric noise. Note, however, for deep neural nets, constant aleatoric noise is not a common assumption, and indeed the main benefit of using BALD over entropy is that it performs well when aleatoric uncertainty varies across samples because it estimates epistemic uncertainty and not aleatoric uncertainty \citep{smith2018understanding}. Moreover, we examine the relationship between BALD and \EPIGBALD in detail and reach a more varied conclusion than \citet{mackay1992information} by specifically taking into account outliers and distribution shift.

In a similar vein, \citet{Roy2001TowardOA} introduce the expected error reduction as acquisition function for individual acquisition in classification tasks without a Bayesian formalism.
Their method corresponds to a non-Bayesian version of \citet{mackay1992information}'s Mean Marginal Information Gain and \EPIG: it essentially computes the expected reduction in generalization loss, \(\Hof{\Yeval \given \Xeval, \Yacqset, \xacqset, \Dtrain}\), which we have examined in \S\ref{sec:epig} as being equivalent to \EPIG.
This requires retraining the model for each possible label of each evaluation sample which they find to be computationally challenging. 

The predictive information as mutual information between the past and future was introduced by \citet{bialek1999predictive} and has been used in reinforcement learning to increase sample efficiency \citep{lee2020predictive}.

\citet{Wang2021BeyondMU} extend the Mean Marginal Information Gain for regression following insights from BatchBALD to the batch setting and evaluate it in a transductive active learning setting \citep{yu2006active}. They focus on measuring the quality of posterior predictive correlations. However, both \citet{yu2006active} and \citet{Wang2021BeyondMU} only examine the case where pool and test-time input distribution are identical.

Unlike much active learning literature whose experiment setting implicitly assumes that that pool and test samples are drawn from the same distribution, \EPIG and \EPIGBALD support the setting in which pool and test distribution do \emph{not} match. While BALD ignores the underlying distribution in general, this is not the case for other diversity-based active learning methods like BADGE \citep{ash2019deep}, for example, which implicitly use the empirical pool set distribution to select diverse samples via clustering.
In contrast, we allow that the unlabeled pool samples might follow a potentially different distribution than the distribution used for evaluation
\begin{math}
 \ppool{x} \not= \ptest{x},
\end{math}
and examine such settings as well.

Finally, \citet{Kothawade2021SIMILARSI} examine general submodular acquisition functions in a non-Bayesian active learning setting. However, they assume access to out-of-distribution data in the distribution shift setting.

To our knowledge, within Bayesian Deep Active Learning, using evaluation samples to guide acquisition in the form of \EPIG and \EPIGBALD, respectively, in settings in which the test distribution and pool distribution are not identical have not been examined, previously.

\begin{fullpaper}
    Within active learning, CoreSet Active Learning is a state-of-the-art approach for deterministic models. VAAL is task-agnostic approach that takes into account the distribution of the unlabeled pool set by using an variational auto-encoder in an adversarial mini-max setting to find samples that look as much unlike the training set as possible.    
\end{fullpaper}

\section{Empirical Validation}
\label{sec:experiments}
\begin{figure}[t]
        \centering
        \includegraphics[width=0.8\linewidth]{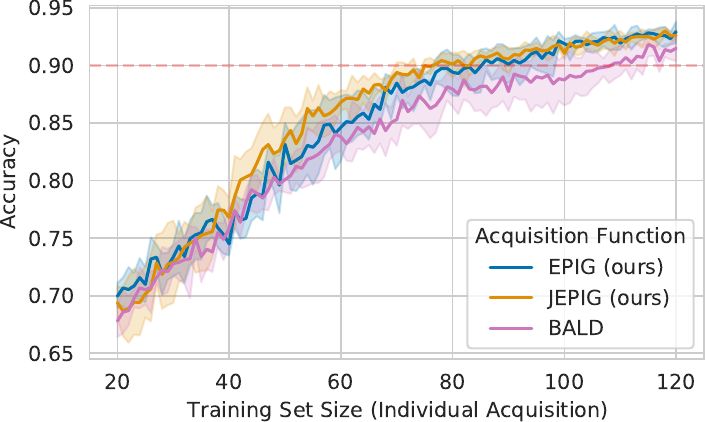}
        \caption{\emph{\EPIG vs \EPIGBALD vs BALD with Bayesian Neural Networks on MNIST.}
        \EPIGBALD performs better under than MC Dropout than \EPIG.
        }
        \label{fig:ablation_epig_vs_epigbald}    
\end{figure}

\begin{figure}[t]
    \centering
    \includegraphics[width=\linewidth]{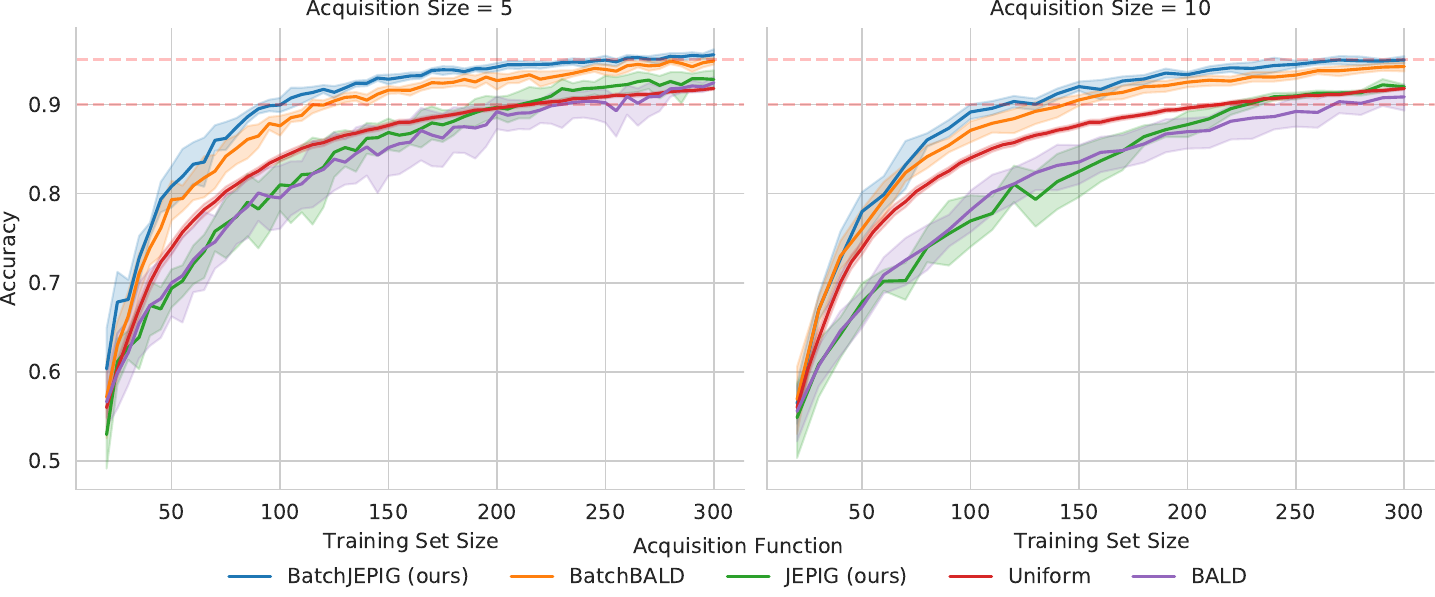}
    \caption{\emph{(Batch)BALD vs (Batch)\EPIGBALD on RepeatedMNIST (MNISTx2).} \EPIGBALD outperforms BALD. %
    }
    \label{fig:normal_active_learning}
\end{figure}

\begin{figure*}[th]
    \centering
    \begin{subfigure}{0.48\linewidth}
        \includegraphics[width=\linewidth]{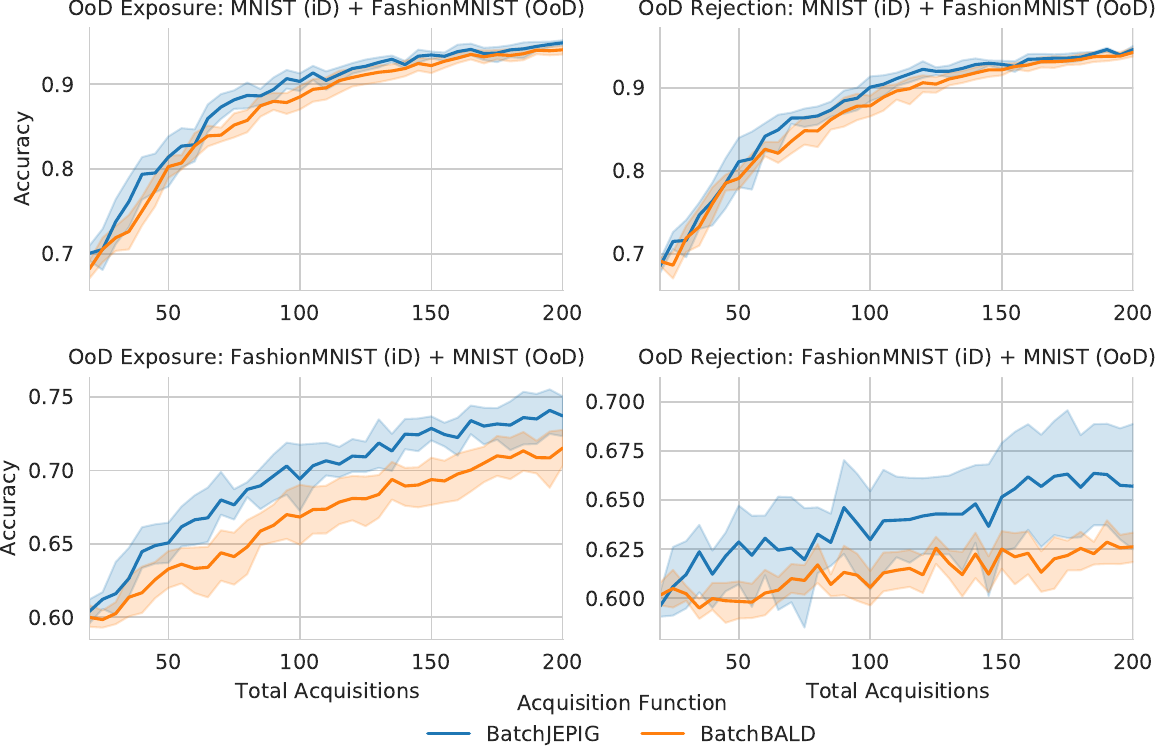}
        \subcaption{Accuracy}
        \label{subfig:ood_mnist_and_fmnist accuracy}
    \end{subfigure}
    \begin{subfigure}{0.48\linewidth}
        \includegraphics[width=\linewidth]{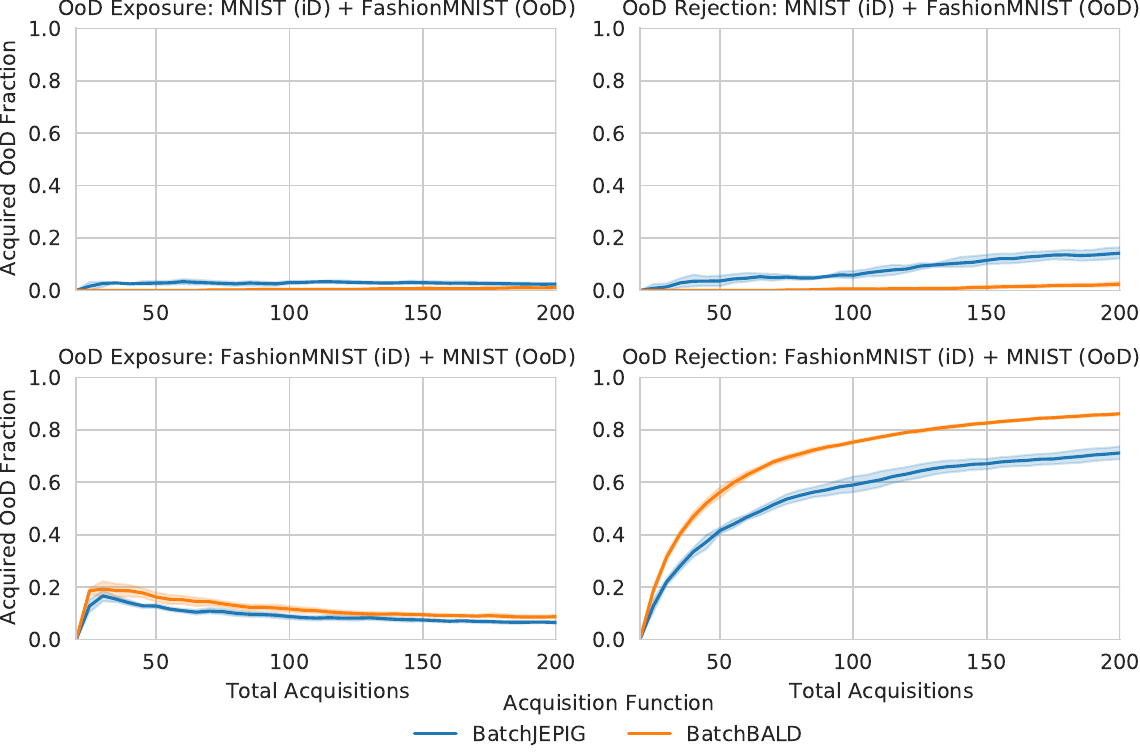}
        \subcaption{Aquired OoD Ratio}
        \label{subfig:ood_mnist_and_fmnist ood fraction}
    \end{subfigure}
    \caption{
        \emph{MNIST and FashionMNIST pairings with OoD rejection or exposure.} \EPIGBALD performs better than BALD. 5 trials.
        }
    \label{fig:ood_mnist_and_fmnist}
\end{figure*}

\begin{figure*}[th]
    \centering
    \begin{subfigure}{0.49\linewidth}
        \includegraphics[width=\linewidth]{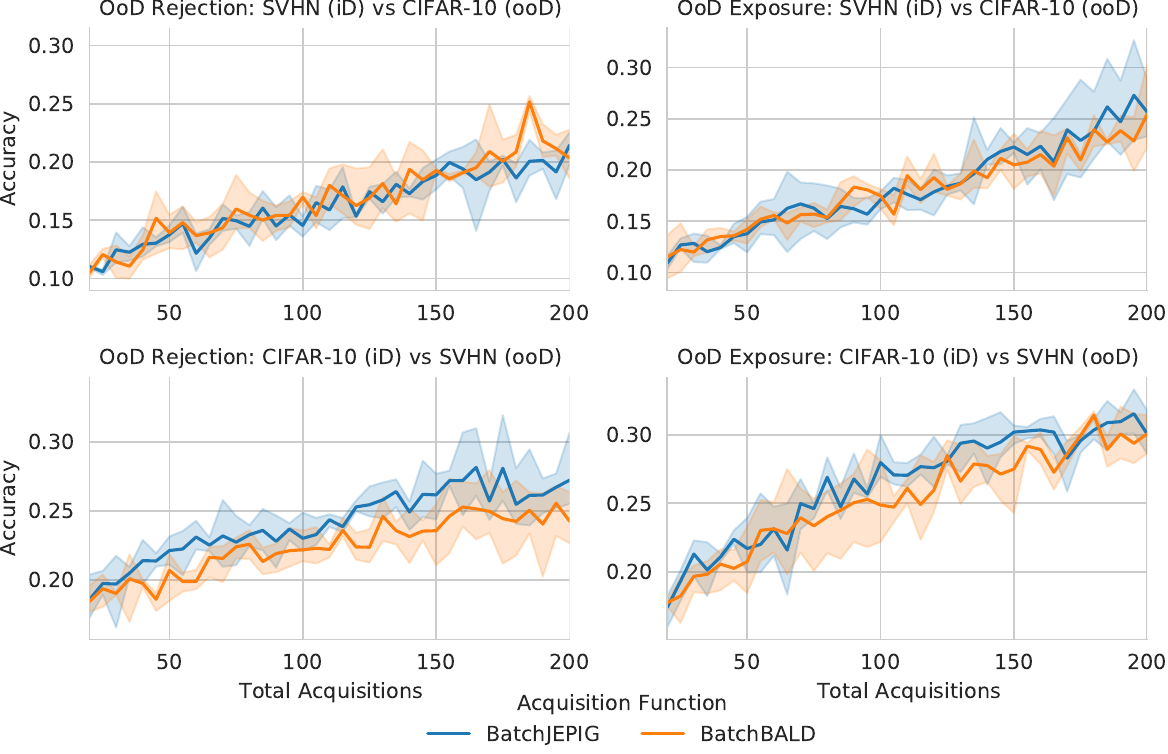}
        \subcaption{Accuracy}
    \end{subfigure}
    \begin{subfigure}{0.49\linewidth}
        \includegraphics[width=\linewidth]{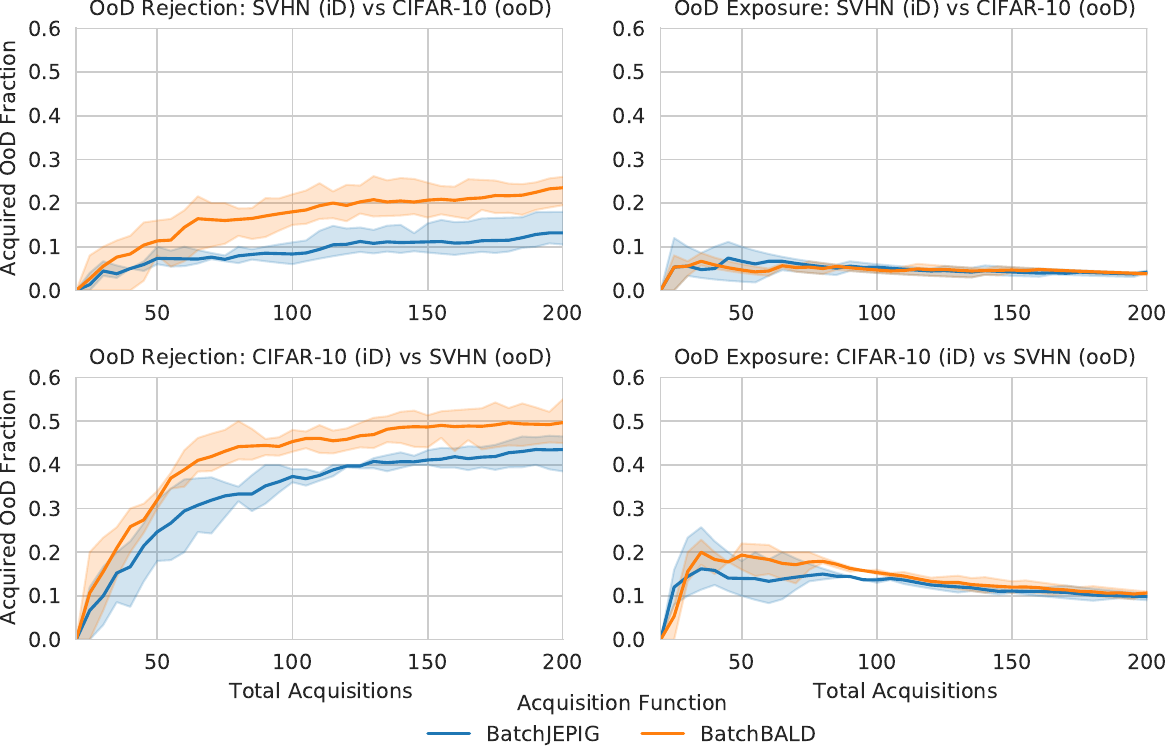}
        \subcaption{Aquired OoD Ratio}
    \end{subfigure}
    \caption{
        \emph{CIFAR-10 and SVHN pairings with OoD rejection or exposure.} \EPIGBALD performs better than BALD. 5 trials. Acquisition size 5. Initial training size 5.
        }
    \label{fig:ood_cifar10_and_svhn}
\end{figure*}

\begin{figure}[th]
    \centering
    \includegraphics[width=\linewidth]{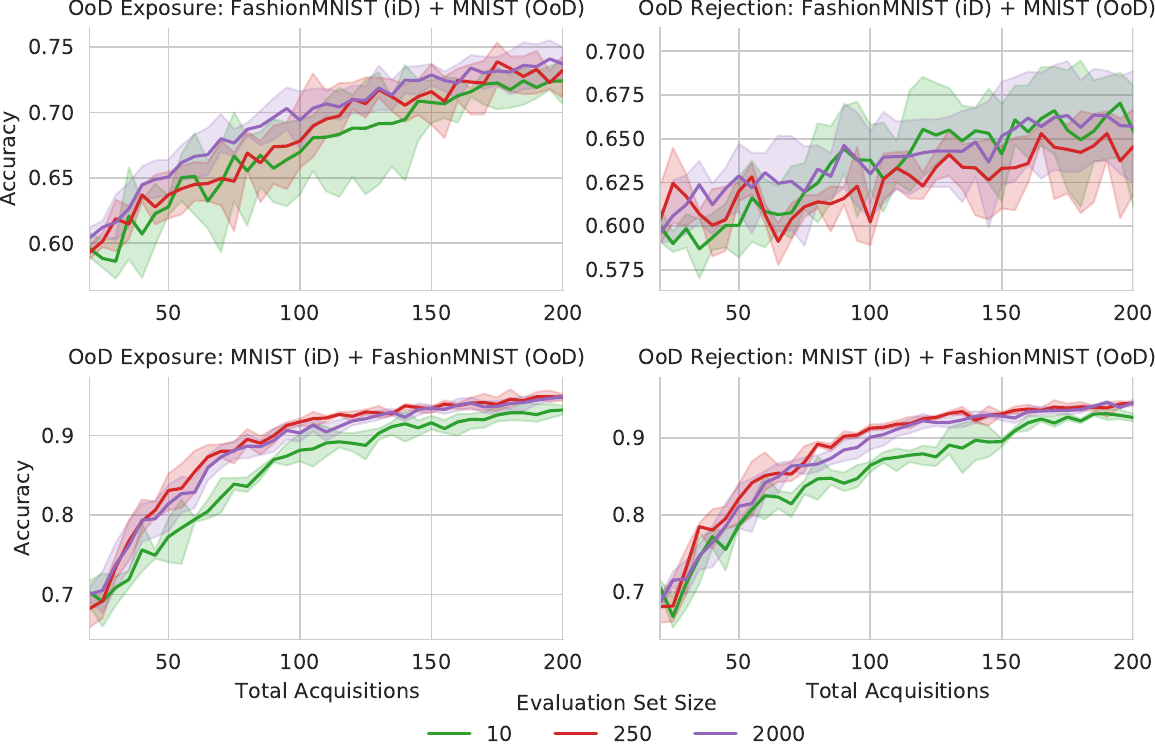}
    \caption{\emph{Evaluation Set Size Ablation}. MNIST and FashionMNIST pairings with OoD rejection or exposure. A larger evaluation set performs better. 5 trials.}
    \label{subfig:ood_mnist_and_fmnist eval_set_size}
\end{figure}

\begin{figure}[th]
    \centering
    \includegraphics[width=0.8\linewidth]{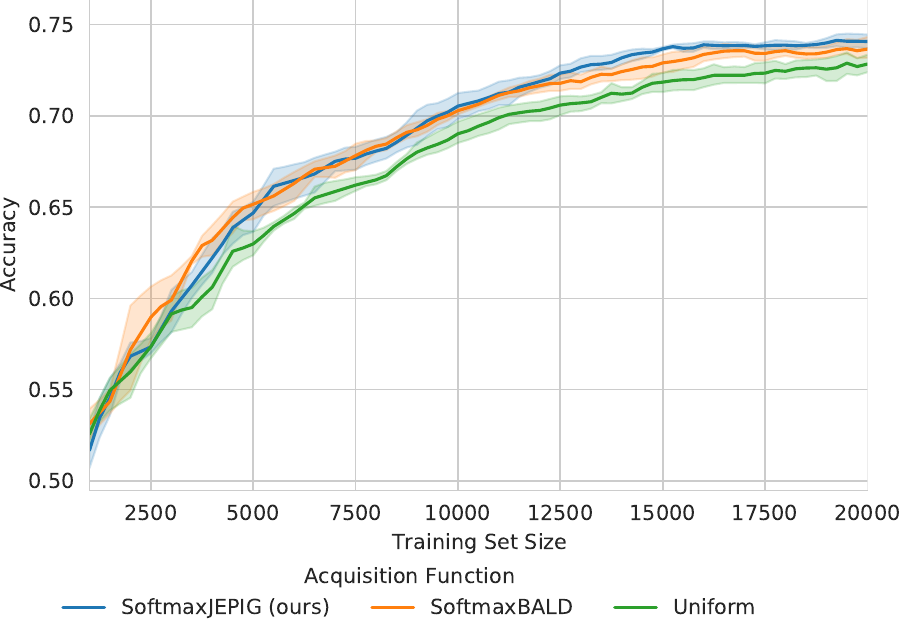}
    \caption{\emph{BALD vs \EPIGBALD on CIFAR-10.} \EPIGBALD outperforms BALD. 5 trials each. With batch acquisition size 250, and initial training size 1000. Median accuracy after smoothing with a Parzen window filter over 30 acquisition steps to denoise.}
    \label{fig:normal_active_learning_cifar10}
\end{figure}

We evaluate the performance of \EPIG and \EPIGBALD using a form of self-distillation described in appendix \S\ref{sec:practical_approximation} in a regular active learning setting and under distribution shift.
We also provide an ablation with different evaluation set sizes. We use approximate BNNs based on MC dropout \citep{gal2016dropout}.

\textbf{Approximate BNNs.}
Exact Bayesian inference is intractable for deep learning, and here we only consider variational inference for approximate inference using a variational distribution $\qof{\w}$. We determine $\qof{\w}$ by minimizing the following KL divergence:
\begin{multline*}
    \Kale{\qof{\w}}{\pof{\w \given \Dtrain}} =\\ 
    = \simpleE{\qof{\w}} [-{\underbrace{\log \pof{\ytrainset \given \xtrainset, \w}}_\text{log likelihood}}] \\
    + \underbrace{\Kale{\qof{\w}}{\pof{\w}}}_\text{prior regularization} + \underbrace{\log \pof{\Dtrain}}_\text{model evidence} \ge 0.
\end{multline*}
We can use the local reparameterization trick and Monte-Carlo (MC) dropout for $\qof{\w}$ in a deep learning context \citep{gal2016dropout}.

\textbf{Setup.} We compare \EPIG and \EPIGBALD with different acquisition sizes to BALD, either using the top-k  of individual scores \citep{gal2016dropout}, the batch version BathBALD \citep{kirsch2019batchbald}---which is equivalent to the previous for individual acquisition---or SoftmaxBALD \citep{powerbald} for larger acquisition batch sizes. SoftmaxBALD samples without replacement from the pool set using the Softmax of the acquisition scores with temperature 8.

On MNIST and MNISTx2 (RepeatedMNIST), we use a LeNet-5 model \citep{lecun1998gradient}, which we train as described in \citet{kirsch2019batchbald}. 

For CIFAR-10 \citep{krizhevsky2009learning}, we use a ResNet18 model \citep{he2016deep} which was modified as described in \citet{kirsch2019batchbald} to add MC dropout to the classifier head and also follows the described training regime. We train with an acquisition batch size of 250 and an intial training set size of 1000.  \citep{powerbald}.
We use MC dropout models with 100 dropout samples when computing the acquisition scores.

\textbf{Performance in Regular Active Learning.}
We evaluate whether ignoring the test--time input distribution has a detrimental effect on BALD even when the pool set distribution matches the test distribution.

For this, we compare BALD and \EPIGBALD in \cref{fig:normal_active_learning} on MNISTx2 and \EPIG and \EPIGBALD in \cref{fig:ablation_epig_vs_epigbald} on MNIST. In the regular active learning setup, there is no distribution shift between the pool set and test set, so we use all of the unlabeled pool set as evaluation set. 

Both in the top-k and the batch variant, \EPIG and \EPIGBALD outperform BALD on MNISTx2 (and also MNIST, not shown). 
On CIFAR-10, \EPIGBALD also outperforms SoftmaxBALD, as depicted in \cref{fig:normal_active_learning_cifar10}.

However, why does \EPIGBALD outperform \EPIG? We hypothesize that this is because \EPIG takes an expectation over the evaluation set using individual points which by itself might be myopic. It might work well for simple models, but taking the whole evaluation set into account and retraining the model might allow for learning better abstraction in deep models, which might not be the case for \EPIG.

\textbf{Performance in Active Learning under Distribution Shift with MNIST and FashionMNIST.}
We want to evaluate how BALD and \EPIGBALD behave under distribution shift, that is when pool set and test distribution do not match.
For this, we add junk out-of-distribution data to the pool set. In this experiment, the pool set contains MNIST and FashionMNIST \citep{xiao2017fashion} while the test set contains one or the other.
We deal with an acquisition function attempting to acquire OoD data in two different modes: \emph{OoD rejection} rejects OoD data from the batch and does not acquire it; while \emph{OoD exposure} acquires OoD data with uniform targets, similar to outlier exposure methods in OoD detection \citep{hendrycks2018deep}. We use an evaluation set with 2000 unlabeled samples.

\EPIGBALD outperforms BALD on in all combinations, see \cref{subfig:ood_mnist_and_fmnist accuracy}. In all cases but one, \EPIGBALD acquires fewer junk/OoD samples, see \cref{subfig:ood_mnist_and_fmnist ood fraction}. The ablation in \cref{subfig:ood_mnist_and_fmnist eval_set_size} shows that larger evaluation sets are beneficial. Note that the evaluation set is unlabeled and thus does not count towards sample acquisitions.

For CIFAR-10 and SVHN \citep{svhn2011}, \EPIGBALD outperforms BALD under distribution shift in all but one combination and selects fewer OoD samples, see \cref{fig:ood_cifar10_and_svhn}. We use an evaluation set with 1000 unlabeled samples.

\begin{fullpaper}
\section{Ablations for the Ladder of Approximations}

\andreas{Talk about other ablations incl. using ensembles instead of self-distillation.}
\end{fullpaper}

\begin{fullpaper}
    \textbf{Monte-Carlo Estimation using Ensembles.}
    A tractable yet expensive approximations of \EPIG and \EPIGBALD can be computed by using Monte-Carlo samples:
    \begin{enumerate}[leftmargin=*]
        \item we sample multiple sets of pseudo-labels $\yevalset \sim \pof{\yevalset \given \xevalset, \Dtrain}$; and
        \item we train an ensemble of models, each with a different set of pseudo-labels $\yevalset$.
    \end{enumerate}
    Evaluating \EPIG or \EPIGBALD using such a Monte-Carlo ensemble will yield an approximation of the true quantities. 
    
    We compare the quality of these approximations in the next section.
\end{fullpaper}

\begin{fullpaper}
\textbf{Ablations.}
We compare self-distillation with an ensemble of models, each trained with a different set of labels for the evaluation set, and with training a model on the true labels of the evaluation set. We also compare evaluation sets of varying sizes: 10, 250, and 1000 unlabeled samples, to determine how much the size affects the performance.

PLACEHOLDER \EPIGBALD strongly outperforms BALD on MNIST, while \EPIG does not. This is because of the approximate Bayesian models separately trained on only the training set and the training set and evaluation set are not compatible. This also shows that the approach stemming from the epistemic uncertainty intuition leads to stable performance. We conclude that the difference in informativeness and epistemic uncertainty (via mutual information) in \EPIGBALD is more meaningful than the difference in overall uncertainty (via conditional entropy) in \EPIG. 
\end{fullpaper}

\FloatBarrier

\section{Conclusion}
We have examined two acquisition functions \EPIG and \EPIGBALD first introduced by \citet{mackay1992information} in a deep learning setting while also drawing new connections to the expected error reduction method in non-Bayesian active learning \citep{Roy2001TowardOA}.
We have highlighted when these methods are of particular importance and why, and we have examined when BALD and \EPIGBALD are equivalent, finding that dismissing \EPIGBALD---as done by \citet{mackay1992information}---might have been premature.
First experiments show promising results both in regular active learning and in active learning under distribution shift.

\section*{Acknowledgements}

The authors would like to thank in particular Tim G. J. Rudner for his feedback at various stages of the project as well as Joost van~Amersfoort and the members of OATML in general. AK is supported by the UK EPSRC CDT in Autonomous Intelligent Machines and Systems (grant reference EP/L015897/1). This work was also supported by the Royal Academy of Engineering under the Research Chair and Senior Research Fellowships scheme, EPSRC/MURI grant EP/N019474/1.

\bibliographystyle{plainnat}
\bibliography{references}
\clearpage
\appendix

\section{Proofs}
\relationjpigepig*
\begin{proof}
    We expand the mutual information to the right:
    \begin{align}
        &\MIof{\Yeval; \Yacqset \given \Xeval, \xacqset, \Dtrain} = \\
        & \quad = \Hof{\Yeval \given \Xeval, \Dtrain} \\
        & \quad \quad - \Hof{\Yeval \given \Xeval, \Yacqset, \xacqset, \Dtrain}.
    \end{align}
    The first term is constant given a fixed evaluation set. The second term is upper-bounded by \(\frac{1}{\lvert \Deval \rvert} \Hof{\Yevalset \given \xevalset, \Yacq, \xacq, \Dtrain}\):
    \begin{align*}
        & \lvert \Deval \rvert \Hof{\Yeval \given \Xeval, \Yacqset, \xacqset, \Dtrain} \notag \\
        & \quad = \sum_i \Hof{\Yeval_i \given \xeval_i, \Yacqset, \xacqset, \Dtrain} \notag \\
        & \quad \ge \Hof{\Yevalset \given \xevalset, \Yacqset, \xacqset, \Dtrain} \notag,
    \end{align*}
    which trivially follows from the chain rule and from the general inequality $\Hof{X} \ge \Hof{X \given Y}$.
    Similarly, we have:
    \begin{align}
        & \MIof{\Yevalset; \Yacqset \given \xevalset, \xacqset, \Dtrain} \\
        & \quad = \Hof{\Yevalset \given \xevalset, \Dtrain} \\
        & \quad \quad - \Hof{\Yevalset \given \xevalset, \Yacqset, \xacqset, \Dtrain},
    \end{align}
    and again the first term is constant given the evaluation set.
    Altogether, the proposition follows from this.
\end{proof}

\section{A Practical Approximation of \(\MIof{\Yacqset ; \W \given \xacqset, \Yevalset, \xevalset, \Dtrain}\)}
\label{sec:practical_approximation}
We want to find an approximation $\hat{\W}$ with distribution $\qof{\hat{\w}}$, such that for all possible acquisition sets, we have:
\begin{align}
    \MIof{\Yacqset ; \W \given \xacqset, \Yevalset, \xevalset, \Dtrain}  \approx \MIof{\Yacqset ; \hat{\W} \given \xacqset}.
\end{align}
We note two properties of this conditional mutual information and the underlying models $\pof{\W \given \yevalset, \xevalset, \Dtrain}$ for different $\yevalset \sim \pof{\yevalset \given \Dtrain}$:
\begin{enumerate}[leftmargin=*]
    \item marginalizing $\pof{\W \given \yevalset, \xevalset, \Dtrain}$ over all possible $\yevalset$ yields the predictions of the original posterior $\pof{\W \given \Dtrain}$, so we would like to have
    \begin{align*}
        \simpleE{\qof{\hat{\w}}}\pof{y \given x, \hat{\w}} = \pof{y \given x, \Dtrain}; 
    \end{align*}
    \item $\MIof{Y ; \W \given x, \Yevalset, \xevalset, \Dtrain} \le \MIof{Y ; \W \given x, \Dtrain}$, and when $x \in \{\xeval_i\}_i$, we expect $\MIof{Y ; \W \given x, \Yevalset, \xevalset, \Dtrain} \ll \MIof{Y ; \W \given x, \Dtrain}$. In other words, the epistemic uncertainty of evaluation samples $\xeval$ ought to decrease when we also train on the evaluation set (using pseudo-labels $\yevalset$), and we would like the same for $\hat{\W}$: 
    \begin{align*}
        \MIof{Y ; \hat{\W} \given x} \le \MIof{Y ; \W \given x, \Dtrain}.
    \end{align*}
\end{enumerate}
Note, that the second property follows from \EPIG being non-negative as a mutual information in two terms and thus so is \EPIGBALD, which means the difference between the two BALD terms is non-negativem and the second property is obtained from that.

Hence, as a tractable approximation $\hat{\W}$, we choose to use a form of \emph{self-distillation}, where we train a model with $\Dtrain$ and the predictions of the original model $\pof{\w \given \Dtrain}$ on $\xevalset$ using a KL-divergence loss, inspired by \citet{hinton2015distilling} and \citet{zhang2019teacher}\footnote{Essentially using \(\alpha=1, \lambda=0\)}. The loss function this is:
\begin{align}
    &L(\xtrainset, \pof{\w}, \qof{\hat{w} \given \Dtrain}) = \notag \\
    & \quad = \frac{1}{\lvert \Dtrain \rvert} \sum_i \Kale{\pof{Y\given \xtrain_i, \Dtrain}}{\qof{Y \given \xtrain_i}} \notag \\
    & \quad \quad + \Kale{\qof{\w}}{\pof{\w}},
\end{align}
with \(\pof{Y\given \xtrain_i} = \simpleE{\pof{\w \given\Dtrain}}{\pof{Y\given \xtrain_i, \w}}\) and \(\qof{Y\given \xtrain_i} = \simpleE{\qof{\hat{\w}}}{\pof{Y\given \xtrain_i, \hat{\w}}}\).
The resulting model posterior $\hat{\W}$ fulfills both properties described above. This is similar to self-distillation in that we train a new model on the predictions of the original model. However, self-distillation does not use predictions on otherwise unlabelled data.
It is also similar to semi-supervised learning \citep{Lee2013PseudoLabelT, yarowsky1995unsupervised} in that we use the predictions of the model on unlabelled data to train a new model. However, semi-supervised learning only uses the samples for which the model is most confident via confidence thresholding and either temperature-scales them for training (soft pseudo-labels) or takes the argmax (hard pseudo-labels) whereas we use the predictions without change for all evaluation samples.

\textbf{Advantages of \EPIGBALD.} Compared to \EPIGBALD, when evaluating
\begin{align}
    &\MIof{\Yevalset; \Yacqset \given \xacqset, \xevalset, \Dtrain} \\
    & \quad = \Hof{\Yacqset \given \xacqset, \Dtrain} \\
    & \quad \quad - \Hof{\Yacqset \given \xacqset, \Yevalset, \xevalset, \Dtrain} \notag \\
    & \quad \approx \varHof{\opq_\emptyset}{\Yacqset \given \xacqset} - \frac{1}{M} \sum_i  \varHof{\opq_{\yevalset^{,i}}}{\Yacqset \given \xacqset}    
\end{align}
with separate approximate Bayesian models \(\opq_\emptyset\) and \(\opq_{\yevalset^{,i}}\) where $\qcof{\emptyset}{\w} \approx \pof{\w \given \Dtrain}$ and $\qcof{\yevalset^{,i}}{\w} \approx \pof{\w \given \yevalset, \xevalset, \Dtrain}$ for $\yevalset^{,i} \sim \pof{\yevalset \given \xevalset, \Dtrain}, i \in 1..M$ for $M$ draws of pseudo-labels for \(\yevalset\), we found that:
\begin{equation*}
    \xHof{\qcof{\emptyset}{Y \given x, \W}} {\color{red}\not=} \frac{1}{M} \sum_i \xHof{\qcof{\yevalset^{,i}}{Y \given x, \W}},
\end{equation*}
which violates the modelling assumption
\begin{equation*}
    \Hof{Y \given x, \W, \Dtrain} = \Hof{Y \given x, \W, \Yevalset, \xevalset, \Dtrain}
\end{equation*}
as $Y \independent \Yevalset \given x, \xevalset, \W$. This is even more of an issue when using a single approximate model with the self-distillation described in the previous section because the two properties we wish for will force \(\Hof{Y \given x, \hat{\W}} \not= \Hof{Y \given x, \Dtrain}\).
This follows immediately from the expansion of the mutual information:
\begin{math}
    \MIof{Y ; \W \given x, \Dtrain}  = \Hof{Y \given x, \Dtrain} - \Hof{Y \given x, \W, \Dtrain}
\end{math}
as the first property will fix \(\Hof{Y \given x, \Dtrain}\) and the second will force the \(\Hof{Y \given x, \W, \Dtrain}\) terms apart to achieve the inequality.
However, \EPIGBALD does not need this assumption as it explicitly estimates the epistemic uncertainty, and thus performs better when using our approximation with self-distillation. Moreover, it has its own strong intuitive motivation.

}

\end{document}